\documentclass{article}

\usepackage{microtype}
\usepackage{graphicx}
\usepackage{booktabs} 
\usepackage{enumitem}
\usepackage{amssymb}
\usepackage{multirow}
\usepackage{multicol}
\usepackage{adjustbox}
\usepackage{colortbl}
\usepackage{caption}
\usepackage{subcaption}
\usepackage{makecell}
\usepackage{pgfplots}
\pgfplotsset{compat=1.18}

\usepackage{hyperref}

\newcommand{\Red}[1]{{\color{red}#1}}

\newcommand{\tb}[1]{{\textbf{#1}}}

\usepackage[accepted]{icml2025}

\usepackage{amsmath}
\usepackage{amssymb}
\usepackage{mathtools}
\usepackage{amsthm}
\usepackage{enumerate}

\usepackage[capitalize,noabbrev]{cleveref}

\theoremstyle{plain}

\theoremstyle{definition}

\theoremstyle{remark}

\usepackage[textsize=tiny]{todonotes}

\icmltitlerunning{Unleashing Uncertainty: Efficient Machine Unlearning for Generative AI}

\begin{document}

\twocolumn[
\icmltitle{Unleashing Uncertainty: Efficient Machine Unlearning for Generative AI}



\icmlsetsymbol{equal}{*}

\begin{icmlauthorlist}
\icmlauthor{Christoforos N. Spartalis}{iti,uva}
\icmlauthor{Theodoros Semertzidis}{iti}
\icmlauthor{Petros Daras}{iti}
\icmlauthor{Efstratios Gavves}{uva,arch}
\end{icmlauthorlist}

\icmlaffiliation{iti}{ITI, Centre for Research \& Technology Hellas, Greece}
\icmlaffiliation{uva}{University of Amsterdam, Netherlands}
\icmlaffiliation{arch}{Archimedes/Athena RC, Greece}

\icmlcorrespondingauthor{Ch. Spartalis}{c.spartalis@uva.nl}

\icmlkeywords{Machine Learning, ICML, Machine Unlearning}

\vskip 0.3in
]



\printAffiliationsAndNotice{}  

\begin{abstract}
We introduce SAFEMax, a novel method for Machine Unlearning in diffusion models. Grounded in information-theoretic principles, SAFEMax maximizes the entropy in generated images, causing the model to generate Gaussian noise when conditioned on impermissible classes by ultimately halting its denoising process.
Also, our method controls the balance between forgetting and retention by selectively focusing on the early diffusion steps, where class-specific information is prominent.
Our results demonstrate the effectiveness of SAFEMax and highlight its substantial efficiency gains over state-of-the-art methods.
\looseness=-1
\end{abstract}
\vspace{-2em}

\section{Introduction}
\label{sec:intro}
Machine Unlearning (MU) for generative AI aims to prevent the generation of impermissible content, such as samples from a specific class—referred to as the \emph{forget samples} or \emph{forget class}. The goal of MU is to correct pre-trained models efficiently, without retraining from scratch, thereby minimizing computational overhead. Efficiency is crucial to reducing both the cost and latency of model correction.
\looseness=-1

Most state-of-the-art methods in generative MU rely heavily on unlearning strategies originally developed for discriminative tasks,
either by directly adapting existing methods to generative settings or by incorporating techniques, such as the Fisher Information Matrix (FIM)~\cite{golatkar2020eternal,foster2024fast} and weight masking~\cite{jia2023model}, 
originally introduced for discriminative MU.
\looseness=-1
 
For example, Selective Amnesia~\cite{heng2023selective} uses FIM to guide unlearning through elastic weight consolidation in generative models.
Saliency Unlearning~\cite{fan2024salun} applies weight masks on top of discriminative MU methods---Random Labeling~\cite{graves2021amnesiac} and Gradient Ascent~\cite{thudi2022unrolling}---which have been extended in generative settings.
Other works~\cite{wu2024munba,wu2024scissorhands,ko2024boosting,patel2025learning} adapt Gradient Ascent or NegGrad+~\cite{kurmanji2023towards} to generative models and apply multi-objective optimization techniques.
\looseness=-1

Despite progress in bridging discriminative and generative MU,
insights from state-of-the-art approaches in discriminative tasks remain underexplored.
Among recent discriminative MU methods, LoTUS~\cite{spartalis2025lotus} stands out for its strong unlearning performance and computational efficiency.
It is an entropy-based method that increases the model's uncertainty on forget samples by smoothing the corresponding prediction probabilities up to an information-theoretic bound, thereby controlling the entropy increase.
Motivated by this approach, we investigate the following research questions in the context of generative MU: 
\looseness=-1

\textbf{1.} Can we efficiently train a generative model to forget specific samples by increasing the entropy on those samples?
\looseness=-1

\textbf{2.} Can we control this process to better balance the trade-off between unlearning and retention of useful knowledge?
\looseness=-1

\begin{figure}
    \centering
    \includegraphics[width=0.71\columnwidth]{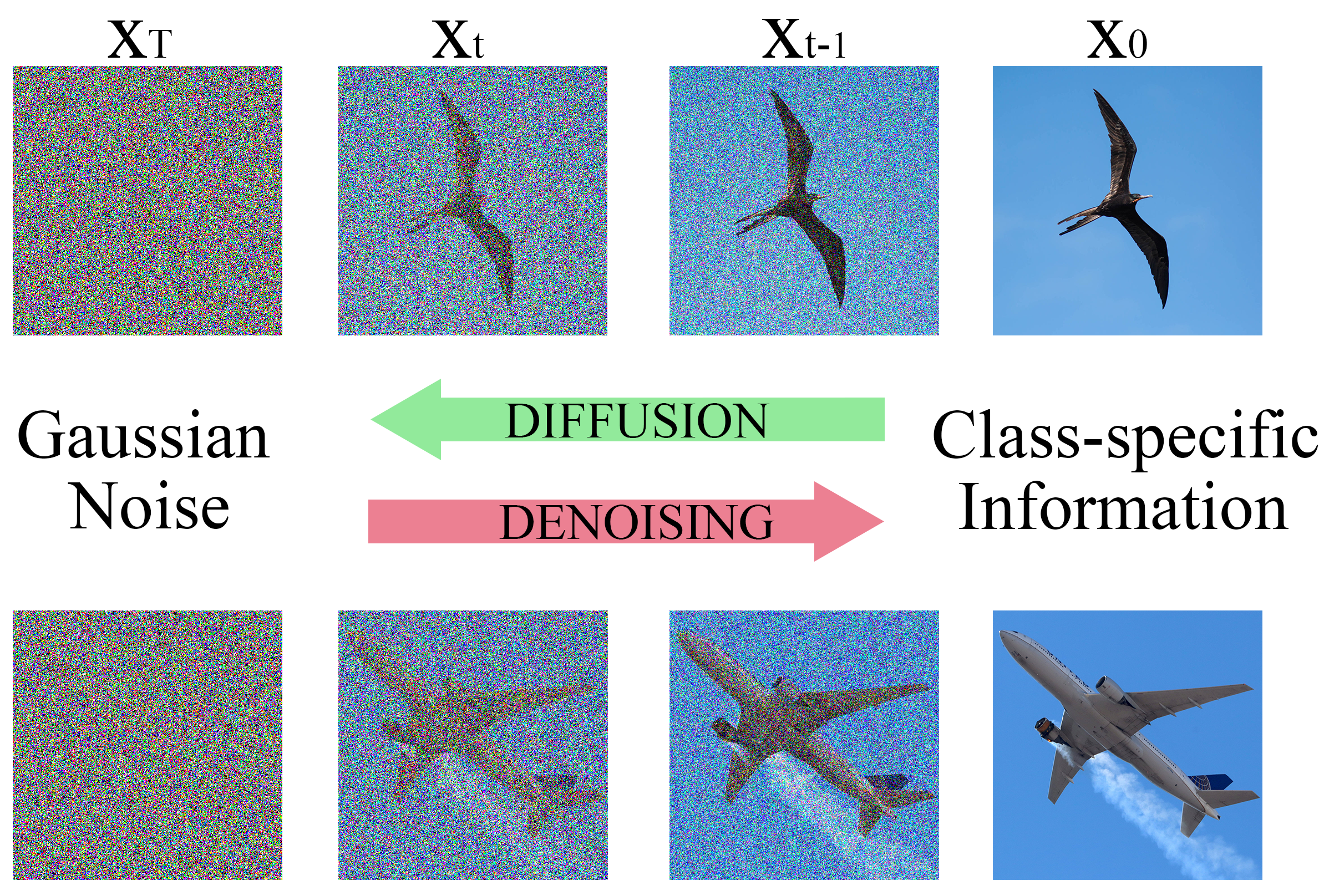}
    \vspace{-0.5em}
\caption{As the diffusion process progresses: (a) the entropy of the latent states increases due to the growing dominance of Gaussian noise, and (b) samples from different classes become increasingly similar. SAFEMax leverages both of these inherent properties of diffusion models to achieve effective and controlled unlearning.\looseness=-1}
\label{fig:noise}
\end{figure}

To this end, we introduce a \textbf{S}imple \textbf{A}nd \textbf{F}ast \textbf{E}ntropy \textbf{Max}imization method for efficient MU in diffusion models.
SAFEMax, for short, maximizes the entropy on the forget samples by training the model to generate Gaussian noise when conditioned on the forget class, ultimately halting the denoising process.
Also, it leverages the progressive loss of information in the diffusion process, as shown in \cref{fig:noise}, to control unlearning and balance forgetting and retention.
Specifically, it applies stronger unlearning in early diffusion steps, where class-specific details are formed and refined, and weaker unlearning in the later steps, where latent states across different classes converge due to the dominant influence of Gaussian noise.
SAFEMax significantly improves efficiency compared to existing unlearning methods while preserving state-of-the-art performance, establishing a scalable and cost-effective solution for MU in generative AI. 
\looseness=-1

\section{Simple and Fast Entropy Maximization for Machine Unlearning in Generative AI}
\label{sec:safemax}
The training of Denoising Diffusion Probabilistic Models (DDPMs)~\cite{ho2020denoising} consists of two processes. First, during the forward diffusion process, the original image $x_0 \!\sim\! q(x)$ is progressively corrupted with Gaussian noise over $T$ steps, such that $x_T \!\sim\! \mathcal{N}(0, \mathbf{I})$.
The noise level at each step is determined by a schedule $\alpha_t$,
and the latent state $x_t$ remains tractable and can be sampled directly:
\looseness=-1
\vspace{-0.5em}
\begin{equation}
q(x_t \mid x_0) = \mathcal{N}(x_t; \sqrt{\bar{\alpha}_t}x_0,\, (1-\bar{\alpha}_t)\mathbf{I}), \quad  \bar{\alpha}_t \!=\! \prod_{i=1}^t\alpha_i
\end{equation}
\looseness=-1
\vspace{-1em}

In the early diffusion steps, the Gaussian noise has lower variance (indicated by $\lim_{t\to0}\bar{\alpha}_t\!=\!1$), making the latent states $x_t$ more similar to the original image $x_0$. These early states contain rich semantic information, which we refer to as \emph{class-specific information}.
As diffusion progresses, the latent states become more entropic (i.e., noisy) because of the dominance of cumulative Gaussian noise.
In later diffusion steps, the variance of the injected Gaussian noise approximates its maximum (indicated by $\lim_{t\to T}\bar{\alpha}_t\!=\!0$) and the distribution $q(x_t)$ becomes increasingly broad.
Consequently, the latent states gradually lose the specific structure of the data and approach a state of maximum entropy. Specifically, for large $T$, the final latent state of the diffusion process, $x_T$, approximates the mean of the training data distribution for all inputs $x_0$~\cite{zhong2024diffusion}.
\looseness=-1

Then, in the denoising process, a denoiser network $\epsilon_\theta$ is trained to predict the noise $\epsilon_t$ at any arbitrary step t:
\looseness=-1
\begin{equation}\label{eq:loss}
    \mathcal{L}= \mathbb{E}_{t\in [1,T],\epsilon \sim \mathcal{N}(0,1)} [\mid\mid \epsilon_t - \epsilon_\theta(x_t, c, t) \mid\mid_2^2]
\end{equation}
where $c$ denotes the class of the original image $x_0$.
\looseness=-1

Unlearning with SAFEMax leverages the inherent Gaussian noise of the diffusion process, particularly the noise $\epsilon_T$,  which corresponds to the maximum entropy latent state $x_T$.
To prevent the denoiser from reconstructing samples of an impermissible class $c_f$, we effectively halt the denoising process and maximize entropy in the model's output by fine-tuning the model with the following \textit{forget loss}:
\begin{equation}\label{eq:forget_loss}
    \mathcal{L}_f = \mathbb{E}_{t\in [1,T],\epsilon \sim \mathcal{N}(0,1)} [\psi(t)\mid\mid \epsilon_T - \epsilon_\theta(x_t, c_f, t) \mid\mid_2^2]
\end{equation}
where $\epsilon_T$ is the cumulative Gaussian noise in the final latent state, $c_f$ is the \textit{forget class}, and $\psi(t)=\exp(-t/T)$ is a monotonically decaying function within the range $[0,1]$ that emphasizes unlearning in the early diffusion steps, where class-specific information is most prominent. This design encourages selective model updates, aiming to better balance unlearning and retention of useful prior knowledge. For the remaining classes, we perform a regular fine-tuning.
\looseness=-1

\paragraph{Information-Theoretic Analysis of MU via Entropy Maximization.}
Consider a training sample $x_0$ and its reconstruction $\hat{x}$ generated by a diffusion model $g(x)$. This process can be viewed as a Markov chain:  $x \mapsto g(x) \mapsto \hat{x}$.
Let $P_e=\text{Pr}\{x \neq \hat{x}\}$ denote the probability of reconstruction error. Within the context of diffusion models, we can define that the equality $x=\hat{x}$ applies when the reconstructed image $\hat{x}$ contains the same semantic information as the original image $x$ (e.g., belonging to the same class).
According to Fano's inequality~\cite{thomas2012elements}, we have:
\looseness=-1
\begin{equation}\label{eq:fano}
   P_e \geq \frac{H(x \mid \hat{x}) - 1}{log\mid\mathcal{X}\mid} 
\end{equation}
where $\mid\mathcal{X}\mid$ is the cardinality of the input space.
\Cref{eq:fano} indicates that as the reconstructed image $\hat{x}$ becomes less informative about the original image $x$ (i.e., as the conditional entropy $H(x \mid \hat{x})$ increases), the lower bound of the error probability $P_e$ increases, implying that the generated image $\hat{x}$ does not share the same semantic information as the original input $x$.
\looseness=-1

To prevent generating images from the \textit{forget class}, we train the model to generate images similar to the ideal latent state $x_T$, which contains pure Gaussian noise (as $T\to\infty$ in the diffusion process) and thus no semantic information from $x$ (denoted as $x_0$ in the diffusion process).
Leveraging the entropy increase that is inherent in the diffusion trajectory, this design maximizes $H(x \mid \hat{x})$, thereby increasing the lower bound of reconstruction error $P_e$ and guiding unlearning.
\looseness=-1

\paragraph{Balancing forgetting and retention with $\psi(t)$.}
Inspired by delineating information for targeted MU in discriminative tasks~\cite{spartalis2025lotus}, and loss scheduling for transfer learning in diffusion models~\cite{zhong2024diffusion}, we introduce a loss scheduling strategy for targeted MU in generative tasks.
Motivated by the observation that class-specific information is most prominent in the early diffusion steps and becomes increasingly obscured by Gaussian noise as the diffusion evolves, as shown in \cref{fig:noise},
we propose a scheduling function that progressively reduces the influence of the forget loss throughout the diffusion process, thereby targeting unlearning in the semantically richer early stages:
\begin{equation}\label{eq:psi}
   \psi(t) = \exp(-\lambda \frac{t}{T}), \quad \text{for} \,\, t \in [0,T] 
\end{equation}
where $\lambda$ is a hyperparameter to controls the rate of decay. 
A larger $\lambda$ causes $\psi(t)$ to decay rapidly, concentrating the unlearning effect more narrowly on the early diffusion steps.
\looseness=-1

\begin{figure*}[]
    \centering
    \begin{minipage}[b]{0.48\textwidth}
        \centering
        \textbf{Unlearning Class 0 (\textit{airplanes})}\par\vspace{0.5em} 

        \begin{subfigure}[b]{0.31\linewidth}
            \centering
            \includegraphics[width=\linewidth]{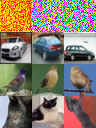}
            \caption{\makecell{SA \\ FID=14.29 \\ UA=98.60\%}}
        \end{subfigure}
        \hfill
        \begin{subfigure}[b]{0.31\linewidth}
            \centering
            \includegraphics[width=\linewidth]{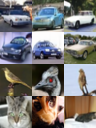}
            \caption{\makecell{SalUn \\ FID=14.11 \\ UA=99.00\%}}
        \end{subfigure}
        \hfill
        \begin{subfigure}[b]{0.31\textwidth}
            \centering
            \includegraphics[width=\linewidth]{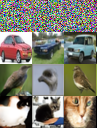}
            \caption{\makecell{SAFEMax (Our) \\ FID=\tb{13.11} \\ UA=\tb{100.00\%}}}
        \end{subfigure}
    \end{minipage}
    \hfill
    \begin{minipage}[b]{0.48\textwidth}
        \centering
        \textbf{Unlearning Class 2 (\textit{birds})}\par\vspace{0.5em} 

        \begin{subfigure}[b]{0.31\textwidth}
            \centering
            \includegraphics[width=\linewidth]{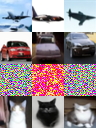}
            \caption{\makecell{SA \\ 18.55 \\ UA=\Red{1.80\%}}}
        \end{subfigure}
        \hfill
        \begin{subfigure}[b]{0.31\textwidth}
            \centering
            \includegraphics[width=\linewidth]{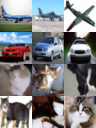}
            \caption{\makecell{SalUn \\ FID=18.24 \\ UA=\tb{98.80\%}}}
        \end{subfigure}
        \hfill
        \begin{subfigure}[b]{0.31\textwidth}
            \centering
            \includegraphics[width=\linewidth]{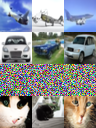}
            \caption{\makecell{SAFEMax (Our) \\ FID=\tb{17.07} \\ UA=\Red{0.00\%}}}
        \end{subfigure}
    \end{minipage}
\vspace{-0.5em}
\caption{\textbf{Qualitative and quantitative results.} SAFEMax generates Gaussian noise for the forget class while preserving high-quality outputs for the retained classes, as reflected in FID. SAFEMax generates noise that is more consistent across forget samples compared to SA, whereas SalUn does not increase entropy and instead replaces forget samples with those of a fixed class. The UA drops for SAFEMax and SA on Class 2 (marked in \textcolor{red}{red}) not due to unlearning failure, but because the classifier erroneously identifies noise as \emph{birds}.}

    \label{fig:qualitative}
\end{figure*}

\begin{table}[ht]
    \centering
    \caption{\textbf{Evaluation of Unlearning} using a ResNet34 classifier on images generated by a DDPM for a forgotten CIFAR-10 class. SAFEMax achieves the highest scores in Unlearning Accuracy (UA) and average entropy ($H$). The class highlighted in \textcolor{red}{red} indicates a case where the classifier failed to evaluate correctly.\looseness=-1}
    \vspace{0.25em}
    \label{tab:clf_eval}
    \begin{adjustbox}{width=1\columnwidth}
        \begin{tabular}{|c|cc|cc|>{\columncolor{blue!7}}c>{\columncolor{blue!7}}c|}
    \hline
        \multirow{2}{*}{\rotatebox[origin=c]{90}{Class}}
        & \multicolumn{2}{c|}{SA~\cite{heng2023selective}}
        & \multicolumn{2}{c|}{SalUn~\cite{fan2024salun}}
        & \multicolumn{2}{c|}{\cellcolor{blue!7}{SAFEMax (Our)}} \\
        & $H \uparrow$ & UA (\%) $\uparrow$
        & $H \uparrow$ & UA (\%) $\uparrow$
        & $H \uparrow$ & UA (\%) $\uparrow$ \\
    \hline
       0 & 1.062      & 98.60 
         & 0.051      & 99.00 
         & \tb{1.132} & \tb{100.00} 
    \\
       1 & 0.987      & 99.60 
         & 0.032      & \tb{100.00}
         & \tb{1.156} & \tb{100.00} 
    \\
       \Red{2}
         & 0.948       & \Red{1.80}
         & 0.084       & \tb{98.80}
         & \tb{1.156}  & \Red{0.00} 
    \\
       3 & 1.006      & \tb{100.00}
         & 0.068      & 99.60
         & \tb{1.122} & \tb{100.00} 
    \\
       4 & 0.926      & \tb{100.00}
         & 0.085      & 99.60
         & \tb{1.128} & \tb{100.00} 
    \\
       5 & 0.908      & \tb{100.00}
         & 0.040      & 99.60
         & \tb{1.118} & \tb{100.00} 
    \\
       6 & 0.993      & \tb{100.00}
         & 0.045      & \tb{100.00}
         & \tb{1.144} & \tb{100.00} 
    \\
       7 & 1.007      & \tb{100.00}
         & 0.027      & \tb{100.00}
         & \tb{1.136} & \tb{100.00} 
    \\
       8 & 0.900      & \tb{100.00}
         & 0.045      & 99.20
         & \tb{1.152} & \tb{100.00} 
    \\
       9 & 0.998      & \tb{100.00}
         & 0.057      & 99.20
         & \tb{1.124} & \tb{100.00} 
    \\
    \hline
    \end{tabular}
    \end{adjustbox}

\end{table}

\begin{table}[ht]
    \centering
\caption{\textbf{Evaluation of Retention \& Efficiency.} We report the mean ($\mu$) and standard deviation ($\sigma$) across all CIFAR-10 classes. SAFEMax achieves the best score in Fr\'echet Inception Distance (FID) on generated images from the non-forgotten classes, Runtime Estimation (RTE) in minutes, and GPU memory usage in GB.\looseness=-1}
\vspace{0.25em}
    \label{tab:fid}
    \begin{adjustbox}{width=1\columnwidth}
        \begin{tabular}{|c|ccc|ccc|>{\columncolor{blue!7}}c>{\columncolor{blue!7}}c>{\columncolor{blue!7}}c|}
    \hline
        \multirow{2}{*}{\rotatebox[origin=c]{90}{Class}}
        & \multicolumn{3}{c|}{SA~\cite{heng2023selective}}
        & \multicolumn{3}{c|}{SalUn~\cite{fan2024salun}}
        & \multicolumn{3}{c|}{\cellcolor{blue!7}{SAFEMax (Our)}} \\
        & FID $\downarrow$ & RTE $\downarrow$ & GPU $\downarrow$ & FID$\downarrow$ & RTE $\downarrow$ & GPU $\downarrow$ & FID$\downarrow$ & RTE $\downarrow$ & GPU $\downarrow$ \\
    \hline
       0 & 14.29      & 174.32    & 17.29
         & 14.11      & 11.56     & 23.22
         & \tb{13.11} & \tb{5.82} & \tb{9.50} 
    \\
       1 & 18.72      & 174.37    & 17.29 
         & \tb{16.85} & 11.96     & 23.23 
         & 18.01      & \tb{5.79} & \tb{9.50}
    \\
       2 & 18.55       & 174.38    & 17.29 
         & 18.24       & 11.97     & 23.24 
         & \tb{17.07}  & \tb{5.80} & \tb{9.50}
    \\
       3 & 17.66      & 174.76    & 17.29 
         & 16.84      & 12.03     & 23.23 
         & \tb{15.64} & \tb{5.89} & \tb{9.50}
    \\
       4 & 17.67      & 174.87    & 17.29 
         & \tb{16.64} & 12.03     & 23.24 
         & 16.89      & \tb{5.80} & \tb{9.50}
    \\
       5 & 17.31      & 174.62    & 17.29 
         & \tb{16.95} & 11.29     & 23.23 
         & 17.07      & \tb{5.79} & \tb{9.50}
    \\
       6 & 17.71      & 173.75    & 17.29 
         & \tb{16.78} & 12.00     & 23.23 
         & 16.80      & \tb{5.79} & \tb{9.50}
    \\
       7 & 18.37      & 173.76    & 17.29 
         & \tb{16.93} & 12.00     & 23.23 
         & 17.93      & \tb{5.90} & \tb{9.50}
    \\
       8 & 18.56      & 174.26    & 17.29 
         & 18.72      & 11.99     & 23.24 
         & \tb{18.20} & \tb{5.80} & \tb{9.50}
    \\
       9 & 18.28      & 174.65    & 17.29 
         & \tb{15.55} & 11.98     & 23.24 
         & 16.66      & \tb{5.85} & \tb{9.50}
    \\
    \hline
       $\mu$ & 17.71 & 174.37 & 17.29
           & 16.76 & 11.81 & 23.23
           & \tb{16.74} & \tb{5.83} & \tb{9.50}
    \\
       $\sigma$ & 1.29 & 0.38 & 0.00
              & 1.27 & 0.25 & 0.01
              & \tb{0.32} & \tb{0.04} & \tb{0.00}
    \\
    \hline
    \end{tabular}
    \end{adjustbox}
\end{table}

\section{Experiments \& Discussion}
\label{sec:discussion}

We follow the experimental setup and evaluation framework of Selective Amnesia (SA) and Saliency Unlearning (SalUn). We aim to forget a specific class within a DDPM trained on CIFAR-10~\cite{krizhevsky2009learning}.
We apply SalUn on top of the Random Labeling strategy using saliency masks that update 50\% of the weights, as proposed by the authors for DDPMs.
For SA, we follow the author's configuration and perform unlearning for 20,000 iterations.
For SAFEMax, we adopt the same hyperparameter values as SalUn and perform unlearning for 1,000 iterations, consistent with SalUn.
\looseness=-1

To evaluate \textbf{unlearning}, we use a ResNet34 classifier, pre-trained on ImageNet~\cite{deng2009imagenet} and fine-tuned on CIFAR-10 for 20 epochs.
We report two key metrics: (i) \emph{Unlearning Accuracy (UA)}, defined as 100\% minus the accuracy of the classifier on the forget class, and (ii) the average \emph{entropy ($H$)} of the classifier's prediction for the forget class, which captures the uncertainty introduced by the unlearning method. To evaluate \textbf{retention} (i.e., the model's performance on the remaining classes), we compute the \emph{Fr\'echet Inception Distance} on generated images for retained classes. Finally, we measure \textbf{efficiency} by reporting each method's \emph{Runtime Estimation (RTE)} and \emph{peak GPU memory usage}.
\looseness=-1

\paragraph{Effectiveness.}
SAFEMax demonstrates strong unlearning performance by excelling in both forgetting and retention of the respective information. Overall, our method achieves better unlearning accuracy (UA) and retention quality (FID) than state-of-the-art approaches, as shown in \cref{tab:clf_eval,tab:fid}.
\looseness=-1

In terms of \textbf{unlearning}, SAFEMax consistently achieves the maximum UA score (100\%) in all-but-one case.
The only exception is Class 2 (\emph{birds}), where the UA drops to 0\%, indicating that the classifier predicts all generated images for the unlearned class as \textit{birds}.
However, this anomaly is not due a failure of  our method.
Instead, the classifier mistakenly identifies Gaussian noise as birds, as evidenced in \cref{fig:qualitative}.
This observation highlights that \textbf{UA scores can be misleading in isolation} and underscores the importance of examining both quantitative and qualitative results.
\looseness=-1

Notably, SAFEMax consistently achieves the highest increase in entropy, directly aligning with its design goal of maximizing uncertainty for the forget class. As shown in \cref{fig:qualitative}, it generates high-entropy noise more reliably than SA, leading to greater classifier uncertainty, as reflected in the entropy scores ($H$) in \cref{tab:clf_eval}.
In contrast, SalUn undermines the entropy-increase objective of Random Labeling in discriminative tasks.
While Random Labeling originally assigned random incorrect labels to forget samples at each unlearning iteration, its adaptation in SalUn maps the forget class to a fixed alternative class (as shown in \cref{fig:qualitative}).
This leads the evaluation model to make incorrect yet high-confidence predictions, as shown in \cref{tab:clf_eval}.
\looseness=-1

In terms of \textbf{retention}---preserving high image quality for the remaining classes as measured by FID---SAFEMax generally achieves state-of-the-art results as shown in \cref{tab:fid}. Overall, SAFEMax delivers effective forgetting and retention while requiring significantly less time and memory.
\looseness=-1

\vspace{-0.5em}
\paragraph{Efficiency.}
As shown in \cref{tab:fid}, a key advantage of SAFEMax is achieving a strong balance between forgetting and retention without relying on computationally intensive techniques (e.g., the use of FIM, weight masks, or regularization terms for multi-objective optimization) that introduce significant overhead before or during unlearning. Compared to SalUn, which runs for the same number of iterations, SAFEMax is $2\times$ faster. Against SA, our method achieves a $30\times$ speed-up due to its ability to unlearn effectively in far fewer iterations. When also accounting for the time to compute the FIM for SA (1226.98 minutes, not included in the RTE metric), SAFEMax offers a total $230\times$ improvement in runtime.
\looseness=-1

In terms of GPU memory usage, SAFEMax is 59\% more efficient than SalUn and 45\% more efficient than SA.
The high memory demands of SalUn and SA result from their reliance on storing the saliency masks and FIM, respectively.
In contrast, SAFEMax does not require complex auxiliary structures to balance forgetting and retention, and instead uses a simple, predefined scheduler.
This simple yet effective design makes SAFEMax not only faster but also more practical for large-scale or resource-constrained applications.
\looseness=-1

\begin{figure}[]
    \begin{subfigure}[b]{0.4\columnwidth}
        \centering
        \caption{\textbf{Effect of decaying scheduler ($\lambda\!=\!1$) vs. no scheduler ($\lambda\!=\!0$)}. SAFEMax improves the image quality for retained classes (see \textbf{5.62\%} improvement in FID), while still unlearning perfectly.\looseness=-1}
        \vspace{0.5em}
        \begin{adjustbox}{width=0.9\textwidth}
        \begin{tabular}{|c|c|c|}
        \hline
        $\lambda$ & {UA (\%) $\uparrow$} & {FID $\downarrow$} \\ \hline
        0 & \cellcolor{green!15} 100.00 & \cellcolor{red!15} 13.89 \\ \hline
        1 & \cellcolor{green!15} 100.00 & \cellcolor{green!15} 13.11 \\
        \hline
        \end{tabular}
        \end{adjustbox}
    \end{subfigure}
    \hfill
    \begin{subfigure}[b]{0.55\columnwidth}
        \centering
        \begin{tikzpicture}
        \begin{axis}[
            width=\columnwidth,
            ymin=79, ymax=105,
            xmin=10, xmax=150, 
            axis lines=middle,
            axis line style={->},
            xlabel={$\,\,\lambda$}, 
            ylabel={UA (\%)},
            xtick={20,70,120},
            ytick={80,90,100},
            xticklabels={1, 50, 100},
            yticklabels={80, 90, 100},
            tick label style={font=\tiny},
            label style={font=\tiny},
            xlabel style={
                at={(axis cs:155,75)}, 
                anchor=mid,
                font=\footnotesize
            },
            ylabel style={
                at={(axis cs:-15,103)}, 
                anchor=south,
                rotate=0,
                font=\fontsize{8pt}{9pt}\selectfont
            },
            xtick align=outside,
            ytick align=outside,
            nodes near coords,
            every node near coord/.style={font=\tiny, anchor=south west},
        ]
        \addplot[
            color=blue,
            mark=*,
            thick,
        ] coordinates {(20,100) (70,98.8) (120,83.8)};
        \end{axis}
        \end{tikzpicture}
        \vspace{-0.5em}
        \caption{\textbf{As $\lambda$ increases, more information is retained}---even for the forget class, as show by the drop in UA.\looseness=-1}
        \label{fig:lineplot}
        \vspace{-1.5em}
    \end{subfigure}

    \vspace{1em}
        
    \begin{subfigure}[b]{0.24\columnwidth}
        \centering
        \includegraphics[width=1\linewidth]{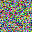}
        \caption{$\lambda\!=\!0$}
        \label{fig:lambda0}
    \end{subfigure}
    \hfill
    \begin{subfigure}[b]{0.24\columnwidth}
        \centering
        \includegraphics[width=1\linewidth]{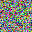}
        \caption{$\lambda\!=\!1$}
        \label{fig:lambda1}
    \end{subfigure}
    \hfill
    \begin{subfigure}[b]{0.24\columnwidth}
        \includegraphics[width=1\linewidth]{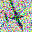}
        \caption{$\lambda\!=\!50$}
        \label{fig:lambda50}
    \end{subfigure}
    \hfill
    \begin{subfigure}[b]{0.24\columnwidth}
    \includegraphics[width=1\linewidth]{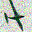}
        \caption{$\lambda\!=\!100$}
        \label{fig:lambda100}
    \end{subfigure}
    \vspace{-0.5em}
\caption{\textbf{Ablation study: Increasing $\lambda$ enhances information retention.} Quantitative and qualitative results from unlearning Class 0 (\textit{airplane}) support our hypothesis behind the decaying scheduler $\psi(t)$: Faster decay (i.e., larger $\lambda$) results in greater information retention, as shown in (b), (e), and (f) and $\psi(t)$ can enhance the quality of generated images from the remaining classes, while still enabling perfect unlearning, as shown in (a), (c), and (d).\looseness=-1}
    \label{fig:ablation}
\end{figure}
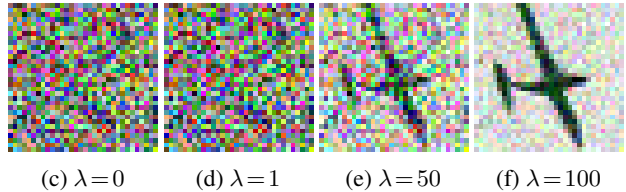
\vspace{-0.5em}
\paragraph{Ablation Study.}
To assess the role of our scheduling function $\psi(t)$, we conducted an ablation study, presented in \cref{fig:ablation}. The results show that SAFEMax achieves strong unlearning performance even without the scheduler, underscoring the robustness of our core method. However, introducing the decaying scheduler $\psi(t)$ improves the trade-off between unlearning and retention, as evidenced by improved image quality in the generated outputs.
We further analyze the impact of the decay parameter $\lambda$ and verify that a faster decay (i.e., larger $\lambda$) improves retention---even for the forget class. In our experiments, we used a moderate value of $\lambda \!=\! 1$, without additional tuning. Tuning $\lambda$ may yield better results.
\looseness=-1

\section{Conclusion}
In this paper, we introduced SAFEMax, an effective and significantly more efficient Machine Unlearning strategy for diffusion models.
Motivated by the information-theoretic analysis of LoTUS~\cite{spartalis2025lotus} in discriminative tasks, we maximized the entropy of the forget samples by leveraging the natural entropy increase inherent to diffusion models.
We further proposed a simple scheduling mechanism that targets class-specific information, enhancing the balance between unlearning and the retention of useful knowledge.
We compared SAFEMax against the most prominent state-of-the-art approaches for DDPM unlearning and showed that our method not only achieves strong unlearning performance but also offers substantial improvements in computational efficiency. These findings suggest that SAFEMax is a promising, scalable, and cost-effective unlearning strategy.
\looseness=-1

\vspace{-1em}
\paragraph{Limitations \& Future Work.}
A more comprehensive evaluation could further strengthen the validity and applicability of our approach, even though SAFEMax has already demonstrated strong performance across key benchmarks.
We plan to compare against more recent DDPM unlearning methods,
and to extend our evaluation to additional datasets and Stable Diffusion models.
\looseness=-1

\section{Acknowledgment}
This work was partially supported by the EU funded project
ATLANTIS (Grant Agreement Number 101073909).


\bibliography{icml2025}
\bibliographystyle{icml2025}

\newpage
\appendix
\onecolumn

\section{Broader Social Impact}
Methods like SAFEMax align machine learning models with privacy regulations and ethical standards, by particularly preventing the generation of impermissible or sensitive content. However, they can also be misused by adversaries to deliberately degrade the performance of otherwise well-functioning models, suggesting careful consideration of deployment practices.
\looseness=-1

\section{Study Note on the Noise Incorporated in SAFEMax and Selective Amnesia}
In Selective Amnesia, the per-pixel noise is drawn from a uniform distribution $\mathcal{U}(-1, 1)$, which has a differential entropy of $H = \log(2) \approx 0.6931$ per dimension (i.e., channel).  
In contrast, in SAFEMax, the per-pixel noise is sampled from a Gaussian distribution $\mathcal{N}(0, 1)$, which has a higher differential entropy of $H = \frac{1}{2} \log(2\pi e) \approx 1.4189$ per dimension.

\section{More Visualizations}
\begin{figure}[h!]
    \centering
    \begin{minipage}[b]{0.32\textwidth}
        \centering
        \textbf{Selective Amnesia} 
        \includegraphics[width=\linewidth]{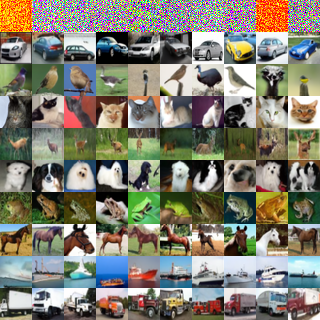}
    \end{minipage}
    \hfill
    \begin{minipage}[b]{0.32\textwidth}
        \centering
        \textbf{Saliency Unlearning}
        \includegraphics[width=\linewidth]{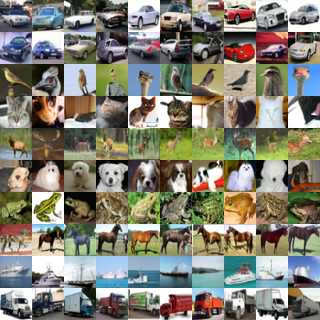}
    \end{minipage} 
    \hfill
    \begin{minipage}[b]{0.32\textwidth}
        \centering
        \textbf{SAFEMax} ($\lambda=1$)
        \includegraphics[width=\linewidth]{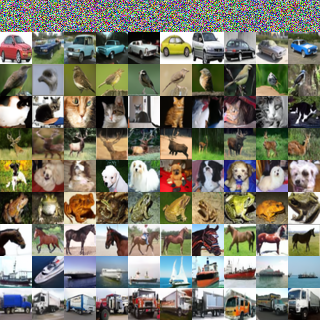}
    \end{minipage}
      
    \centering
    \begin{minipage}[b]{0.32\textwidth}
        \centering
        \textbf{SAFEMax} ($\lambda=0$)
        \includegraphics[width=\linewidth]{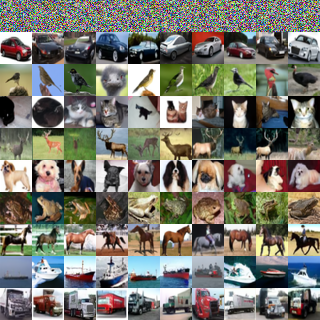}
    \end{minipage}
    \hfill
    \begin{minipage}[b]{0.32\textwidth}
        \centering
        \textbf{SAFEMax} ($\lambda=50$)
        \includegraphics[width=\linewidth]{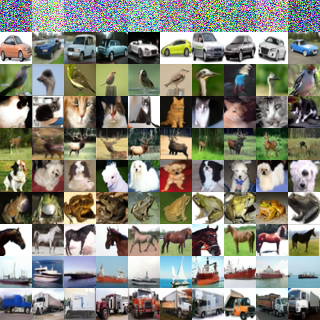}
    \end{minipage} 
    \hfill
    \begin{minipage}[b]{0.32\textwidth}
        \centering
        \textbf{SAFEMax} ($\lambda=100$)
        \includegraphics[width=\linewidth]{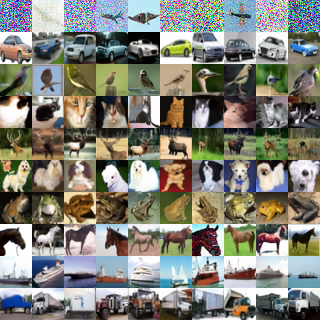}
    \end{minipage}
\vspace{-1em}
\caption{Unlearning Class 0 (\textit{airplanes}).}
\end{figure}

\begin{figure}[h!]
    \centering
    \begin{minipage}[b]{0.32\textwidth}
        \centering
        \textbf{Selective Amnesia} 
        \includegraphics[width=\linewidth]{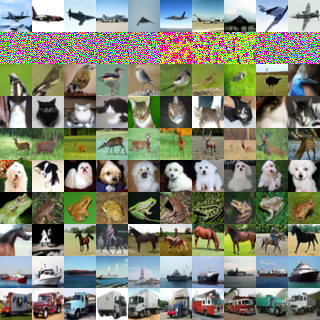}
    \end{minipage}
    \hfill
    \begin{minipage}[b]{0.32\textwidth}
        \centering
        \textbf{Saliency Unlearning}
        \includegraphics[width=\linewidth]{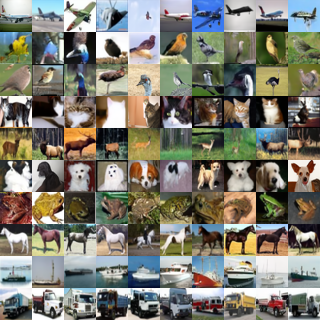}
    \end{minipage} 
    \hfill
    \begin{minipage}[b]{0.32\textwidth}
        \centering
        \textbf{SAFEMax} ($\lambda=1$)
        \includegraphics[width=\linewidth]{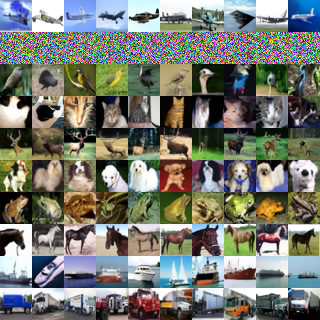}
    \end{minipage}
\vspace{-1em}
\caption{Unlearning Class 1 (\textit{cars}).}
\end{figure}

\begin{figure}[h!]
    \centering
    \begin{minipage}[b]{0.32\textwidth}
        \centering
        \textbf{Selective Amnesia} 
        \includegraphics[width=\linewidth]{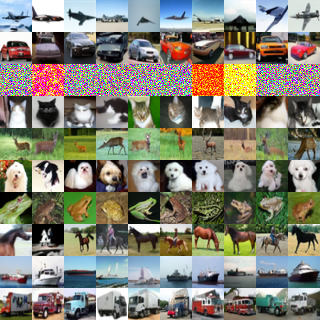}
    \end{minipage}
    \hfill
    \begin{minipage}[b]{0.32\textwidth}
        \centering
        \textbf{Saliency Unlearning}
        \includegraphics[width=\linewidth]{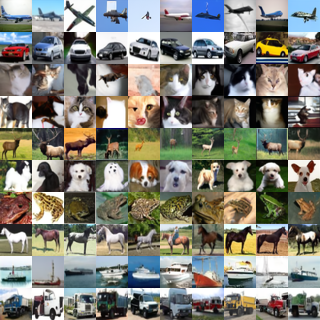}
    \end{minipage} 
    \hfill
    \begin{minipage}[b]{0.32\textwidth}
        \centering
        \textbf{SAFEMax} ($\lambda=1$)
        \includegraphics[width=\linewidth]{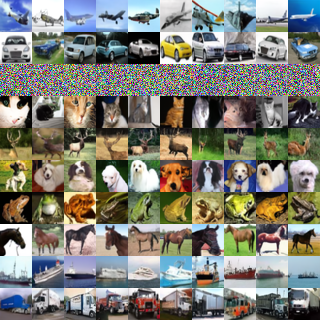}
    \end{minipage}
\vspace{-1em}
\caption{Unlearning Class 2 (\textit{birds}).}
\end{figure}

\begin{figure}[h!]
    \centering
    \begin{minipage}[b]{0.32\textwidth}
        \centering
        \textbf{Selective Amnesia} 
        \includegraphics[width=\linewidth]{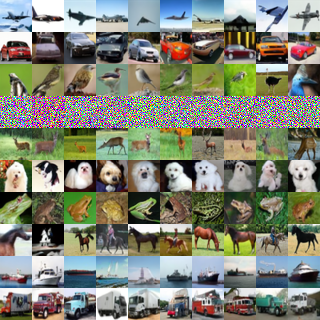}
    \end{minipage}
    \hfill
    \begin{minipage}[b]{0.32\textwidth}
        \centering
        \textbf{Saliency Unlearning}
        \includegraphics[width=\linewidth]{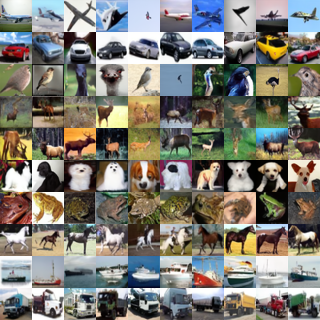}
    \end{minipage} 
    \hfill
    \begin{minipage}[b]{0.32\textwidth}
        \centering
        \textbf{SAFEMax} ($\lambda=1$)
        \includegraphics[width=\linewidth]{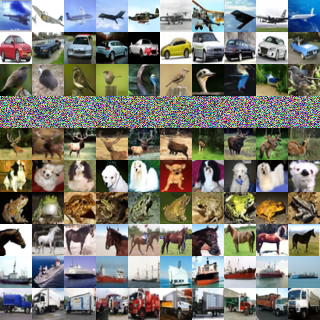}
    \end{minipage}
\vspace{-1em}
\caption{Unlearning Class 3 (\textit{cats}).}
\end{figure}

\begin{figure}[h!]
    \centering
    \begin{minipage}[b]{0.32\textwidth}
        \centering
        \textbf{Selective Amnesia} 
        \includegraphics[width=\linewidth]{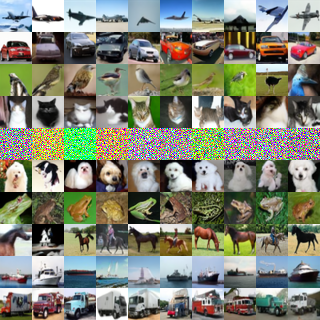}
    \end{minipage}
    \hfill
    \begin{minipage}[b]{0.32\textwidth}
        \centering
        \textbf{Saliency Unlearning}
        \includegraphics[width=\linewidth]{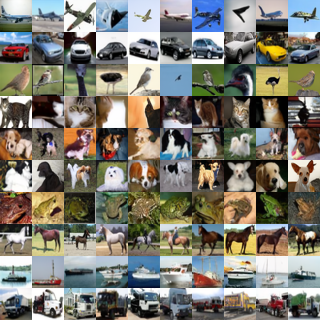}
    \end{minipage} 
    \hfill
    \begin{minipage}[b]{0.32\textwidth}
        \centering
        \textbf{SAFEMax} ($\lambda=1$)
        \includegraphics[width=\linewidth]{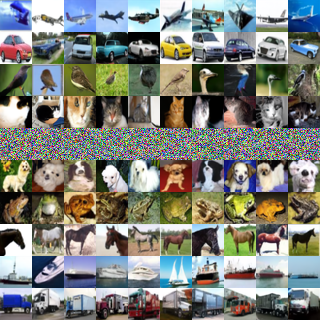}
    \end{minipage}
\vspace{-1em}
\caption{Unlearning Class 4 (\textit{deer}).}
\end{figure}

\begin{figure}[h!]
    \centering
    \begin{minipage}[b]{0.32\textwidth}
        \centering
        \textbf{Selective Amnesia} 
        \includegraphics[width=\linewidth]{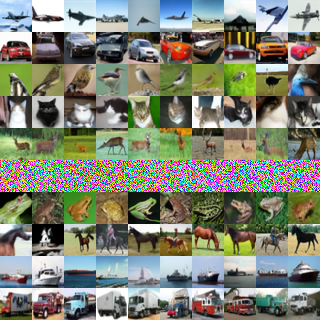}
    \end{minipage}
    \hfill
    \begin{minipage}[b]{0.32\textwidth}
        \centering
        \textbf{Saliency Unlearning}
        \includegraphics[width=\linewidth]{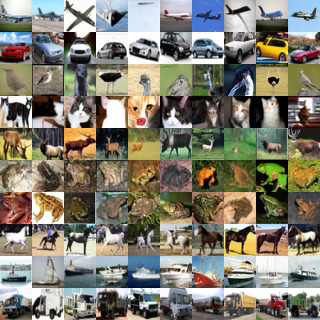}
    \end{minipage} 
    \hfill
    \begin{minipage}[b]{0.32\textwidth}
        \centering
        \textbf{SAFEMax} ($\lambda=1$)
        \includegraphics[width=\linewidth]{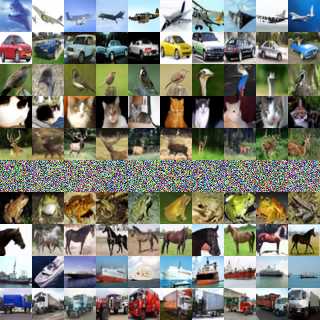}
    \end{minipage}
\vspace{-1em}
\caption{Unlearning Class 5 (\textit{dogs}).}
\end{figure}

\begin{figure}[h!]
    \centering
    \begin{minipage}[b]{0.32\textwidth}
        \centering
        \textbf{Selective Amnesia} 
        \includegraphics[width=\linewidth]{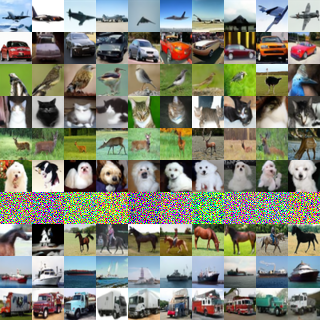}
    \end{minipage}
    \hfill
    \begin{minipage}[b]{0.32\textwidth}
        \centering
        \textbf{Saliency Unlearning}
        \includegraphics[width=\linewidth]{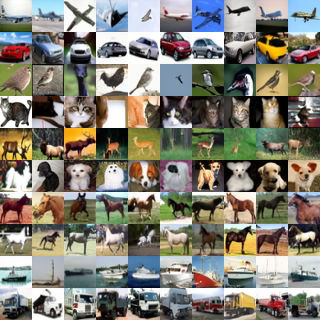}
    \end{minipage} 
    \hfill
    \begin{minipage}[b]{0.32\textwidth}
        \centering
        \textbf{SAFEMax} ($\lambda=1$)
        \includegraphics[width=\linewidth]{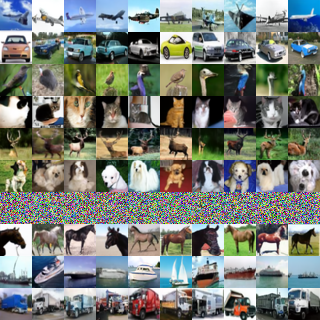}
    \end{minipage}
\vspace{-1em}
\caption{Unlearning Class 6 (\textit{frogs}).}
\end{figure}

\begin{figure}[h!]
    \centering
    \begin{minipage}[b]{0.32\textwidth}
        \centering
        \textbf{Selective Amnesia} 
        \includegraphics[width=\linewidth]{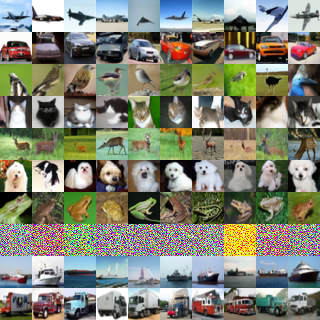}
    \end{minipage}
    \hfill
    \begin{minipage}[b]{0.32\textwidth}
        \centering
        \textbf{Saliency Unlearning}
        \includegraphics[width=\linewidth]{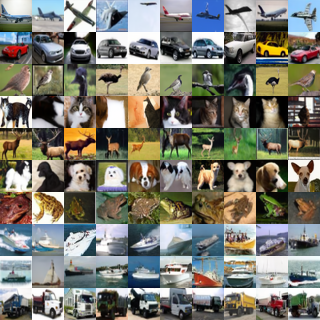}
    \end{minipage} 
    \hfill
    \begin{minipage}[b]{0.32\textwidth}
        \centering
        \textbf{SAFEMax} ($\lambda=1$)
        \includegraphics[width=\linewidth]{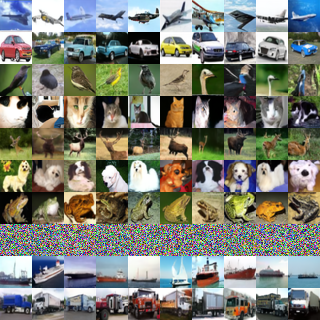}
    \end{minipage}
\vspace{-1em}
\caption{Unlearning Class 7 (\textit{horses}).}
\end{figure}

\begin{figure}[h!]
    \centering
    \begin{minipage}[b]{0.32\textwidth}
        \centering
        \textbf{Selective Amnesia} 
        \includegraphics[width=\linewidth]{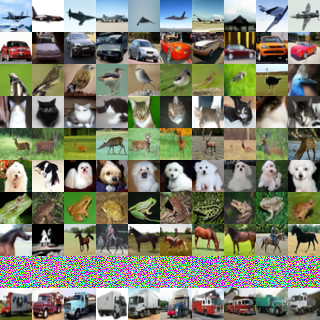}
    \end{minipage}
    \hfill
    \begin{minipage}[b]{0.32\textwidth}
        \centering
        \textbf{Saliency Unlearning}
        \includegraphics[width=\linewidth]{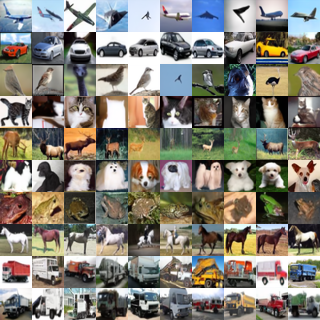}
    \end{minipage} 
    \hfill
    \begin{minipage}[b]{0.32\textwidth}
        \centering
        \textbf{SAFEMax} ($\lambda=1$)
        \includegraphics[width=\linewidth]{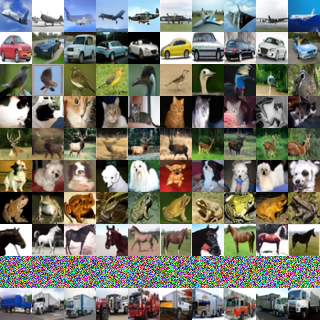}
    \end{minipage}
\vspace{-1em}
\caption{Unlearning Class 8 (\textit{ships}).}
\end{figure}

\begin{figure}[h!]
    \centering
    \begin{minipage}[b]{0.32\textwidth}
        \centering
        \textbf{Selective Amnesia} 
        \includegraphics[width=\linewidth]{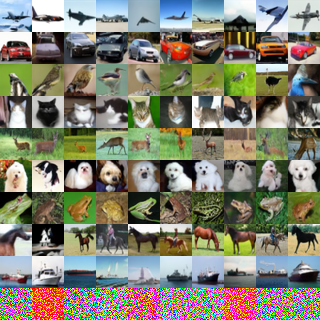}
    \end{minipage}
    \hfill
    \begin{minipage}[b]{0.32\textwidth}
        \centering
        \textbf{Saliency Unlearning}
        \includegraphics[width=\linewidth]{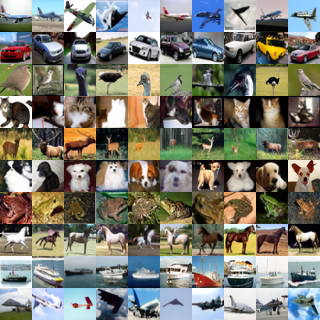}
    \end{minipage} 
    \hfill
    \begin{minipage}[b]{0.32\textwidth}
        \centering
        \textbf{SAFEMax} ($\lambda=1$)
        \includegraphics[width=\linewidth]{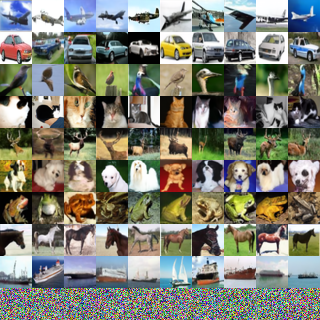}
    \end{minipage}
\vspace{-1em}
\caption{Unlearning Class 9 (\textit{trucks}).}
\end{figure}



\end{document}